# Self-supervised Learning of Rotation-invariant 3D Point Set Features using Transformer and its Self-distillation

Takahiko Furuya [a],1, Zhoujie Chen [a,b], Ryutarou Ohbuchi [a], Zhenzhong Kuang [b]

[a]*Department of Computer Science and Engineering, University of Yamanashi, 4-3-11 Takeda, Kofu-shi, Yamanashi-ken, 400-8511, Japan*
[b]*School of Computer Science, Hangzhou Dianzi University, Hangzhou 310000, China*

**Abstract**

Invariance against rotations of 3D objects is an important property in analyzing 3D point set data. Conventional 3D point set DNNs having rotation invariance typically obtain accurate 3D shape features via supervised learning by using labeled 3D point sets as training samples. However, due to the rapid increase in 3D point set data and the high cost of labeling, a framework to learn rotation-invariant 3D shape features from numerous unlabeled 3D point sets is required. This paper proposes a novel self-supervised learning framework for acquiring accurate and rotation-invariant 3D point set features at object-level. Our proposed lightweight DNN architecture decomposes an input 3D point set into multiple global-scale regions, called tokens, that preserve the spatial layout of partial shapes composing the 3D object. We employ a self-attention mechanism to refine the tokens and aggregate them into an expressive rotation-invariant feature per 3D point set. Our DNN is effectively trained by using pseudo-labels generated by a self-distillation framework. To facilitate the learning of accurate features, we propose to combine multi-crop and cut-mix data augmentation techniques to diversify 3D point sets for training. Through a comprehensive evaluation, we empirically demonstrate that, (1) existing rotation-invariant DNN architectures designed for supervised learning do not necessarily learn accurate 3D shape features under a self-supervised learning scenario, and (2) our proposed algorithm learns rotation-invariant 3D point set features that are more accurate than those learned by existing algorithms. Code is available at: https://github.com/takahikof/RIPT_SDMM

## 1. Introduction

Recent advances in 3D point set analysis technology and prevalence of 3D sensor devices have led to various applications, including autonomous driving of vehicles and robots, maintenance of man-made structures, and terrain inspection for disaster prevention. In 3D point set analysis, 3D shape features extracted from 3D point set data play an important role. To ensure accurate and robust analysis, the 3D shape features should represent high-level semantics of 3D objects and be invariant against geometric transformations including translation, scaling, and rotation of 3D objects. Of these transformations, invariance against rotation, or, more formally, invariance against the 3D rotation group SO(3), is difficult to achieve. Rotation invariance is crucial since the orientations of 3D objects are typically inconsistent. For example, the orientation of a 3D point set obtained by using a 3D range scanner varies depending on the pose and position of the scanned object and the scanner. Or, the correspondence between the upright direction of a 3D object and the coordinate axes of the 3D modeling software is not unique. Therefore, practical 3D point set analysis techniques require 3D shape features that are not influenced by the SO(3) rotations of the input 3D point sets.

In the past few years, several rotation-invariant (RI) deep neural networks (DNNs) for 3D point set analysis have been proposed [1]. These RI DNNs are trained under the supervised learning framework by using labeled point sets as training samples. These studies achieve rotation invariance by, for example, normalizing the rotation of global/local 3D shapes [2][3][4][5] or extracting inherently RI 3D geometric features [6][7][8][9]. Regardless of the methods used to achieve rotation invariance, the existing RI DNNs demonstrated high accuracy as well as rotation invariance for analyzing both realistic 3D point sets obtained by scanning real-world 3D objects and synthetic 3D point sets derived from 3D CAD models.

However, obtaining RI 3D shape features via supervised learning is not always practical due to the high cost of manual annotation to 3D point set data. 3D point sets generated by range scanners are often stored in databases without being labeled. Even if 3D point sets were annotated by humans and collected in online databases, such annotated 3D point sets would be difficult to use directly for supervised learning due to the inconsistency in the human-annotated tags [10]. With these backgrounds, it is crucial to develop a learning framework that leverages a large amount of unlabeled 3D point sets for the acquisition of accurate RI 3D shape features.

Recently, Self-Supervised Learning (SSL) of 3D shape features, which generates supervision signals for DNN training from unlabeled point set data, has attracted attention [11]. However, most existing studies on SSL for 3D point set feature extraction do not consider rotation invariance since their DNN training heavily depends on 3D shapes whose orientations are consistently aligned. SSL of 3D shape features having rotation invariance is barely explored. [12] and [13] are among the few prior studies on SSL of RI 3D point set features. Note, however, that these previous studies aim at SSL of "part-level", or local, RI point set features intended for 3D point set registration. To the best of our knowledge, there exist no previous studies on SSL of "object-level" RI 3D point set features. Object-level features, that represent entire 3D objects such as chairs and cars, can be used for various tasks including retrieval, classification, and clustering of 3D shapes.

Our goal in this paper is to obtain accurate object-level 3D point set features having rotation invariance under the framework of SSL. To achieve this goal, we need to tackle the following two challenges. (1) First, we must design a DNN architecture suitable for our objective. The need to design a new DNN architecture is motivated by the problem of pose information loss highlighted by Chen et al. [8]. This problem arises when local regions of a 3D point set are processed either by rotation normalization or by extraction of inherently RI features. Since such processing discards the orientation information of the local regions, their spatial layout that composes the original entire 3D shape at its object level is lost. Therefore, the existing RI DNN architectures potentially have difficulty in extracting an accurate global shape feature from the set of local features having pose ambiguity. The previous studies have successfully acquired highly semantic RI features with the help of semantic labels. However, in the context of SSL, semantic labels are not available. (2) Second, we have to devise an SSL algorithm to effectively train our DNN. In the field of 2D image analysis, a number of SSL algorithms have been proposed to obtain visual features without semantic labels [30]. In particular, self-distillation with no labels (DINO) [16] is a powerful SSL algorithm built upon a self-distillation framework. Also, research in 2D image analysis has established effective data augmentation techniques, such as CutMix [17] and MultiCrop [18], to facilitate DNN training.

Considering the abovementioned challenges and related studies, we propose a novel DNN architecture and an SSL algorithm tailored for learning RI 3D point set features. Our DNN architecture is called *Rotation-Invariant Point set token Transformer* (*RIPT*). Our SSL algorithm is called *Self-Distillation with Multi-crop and cut-Mix point set augmentation* (*SDMM*).

Our DNN, i.e., RIPT illustrated in Figure 1, first transforms an input 3D point set into multiple "tokens", each of which describes a rotation-normalized 3D shape at a *global* scale. The multiple global tokens per 3D point set are then processed by the self-attention mechanism [14][15] to produce an expressive object-level 3D shape feature. Unlike existing 3D point set DNNs that extract *local* features in shallow layers, RIPT is unique in that its first layer computes global features, each of which covers the entire shape of an input 3D point set. Our RI global feature has a 3D grid structure that can preserve the spatial layout of local regions of the input 3D shape. The use of such grid-structured global feature is expected to avoid the pose information loss problem [8] since there is no pose ambiguity among partial shapes described by the global feature. We design the architecture of RIPT to be computationally efficient since our SSL algorithm requires a large minibatch containing diverse training samples.

Our SSL algorithm, i.e., SDMM depicted in Figure 2, fuses the powerful self-distillation-based SSL algorithm DINO [16] and the powerful data augmentation technique CutMix [17], both originally proposed for 2D image analysis. Identical to DINO, SDMM requires two DNNs with the same architecture (e.g., RIPT), referred to as the student and the teacher, respectively. SDMM learns features via a task in which the student DNN predicts pseudo-labels generated by the teacher DNN. To regularize the training and learn accurate features, SDMM combines the ideas of two data augmentation techniques, namely, MultiCrop [18] incorporated in DINO, and CutMix. Our multi-crop augmentation creates multiple global and local 3D shapes from a training (global) 3D point set. Our cut-mix augmentation mixes parts of two different global 3D point sets and their pseudo-labels. The combination of multi-crop and cut-mix can create minibatches consisting of highly diverse 3D point sets. Note that SDMM is not tied to rotation invariance of the RIPT network architecture. SDMM can potentially be combined with other 3D point set analysis DNNs.

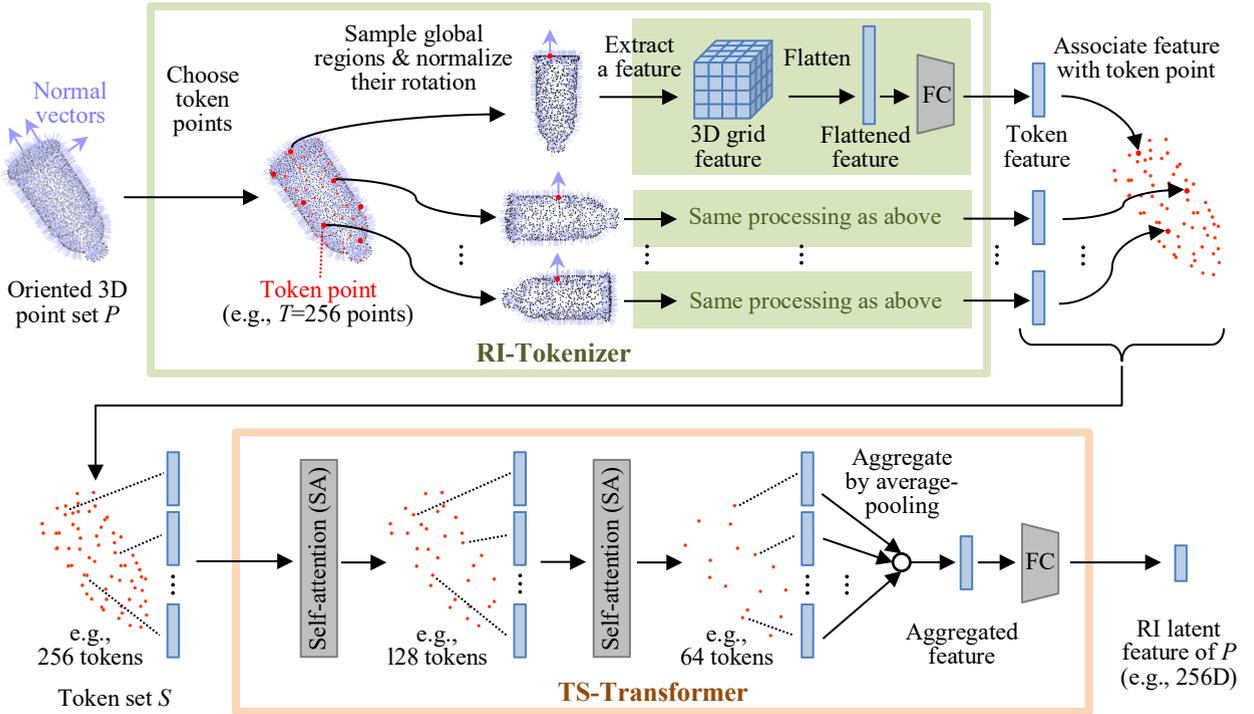

Figure 1. Processing pipeline of the proposed DNN architecture RIPT. RI-Tokenizer transforms the input 3D point set into multiple token features, each of which is rotation-invariant and describes the global 3D geometry of the input. TS-Transformer refines and aggregates the set of tokens to produce the expressive RI latent feature.

We comprehensively evaluate the effectiveness of our proposed framework in terms of both feature accuracy and training efficiency. We found that RIPT trained by SDMM can extract RI 3D point set features that are more accurate than those extracted by using the existing RI DNNs. Most of the existing RI DNNs suffer from low accuracy of their learned 3D shape features when trained under the SSL framework. This paper is the first to identify the incompatibility between the existing RI DNNs and SSL. We also found that the training of RIPT requires less memory and less time than the existing RI DNNs.

The contributions of this paper can be summarized as follows.

- Proposing a novel DNN architecture RIPT suitable for learning RI 3D point set features. RIPT employs the following two strategies: (1) extracting a set of global-scale tokens to encode spatial layouts of local shapes, and (2) refining and aggregating tokens by using the self-attention mechanism to produce an expressive RI object-level feature.

- Proposing a novel SSL algorithm SDMM which trains RIPT under the self-distillation framework. SDMM creates minibatches containing highly diverse training samples by using multi-crop and cut-mix data augmentation techniques.

- Empirically demonstrating rotation invariance, high feature accuracy, and high training efficiency of the proposed algorithm. We also demonstrate, for the first time, that existing RI 3D point set DNNs designed for supervised learning do not necessarily learn accurate 3D point set features under a self-supervised learning scenario.

The rest of this paper is organized as follows. Section 2 reviews related work and Section 3 describes our proposed algorithm. Section 4 discusses the experimental results. Finally, the conclusion and future work are presented in Section 5.

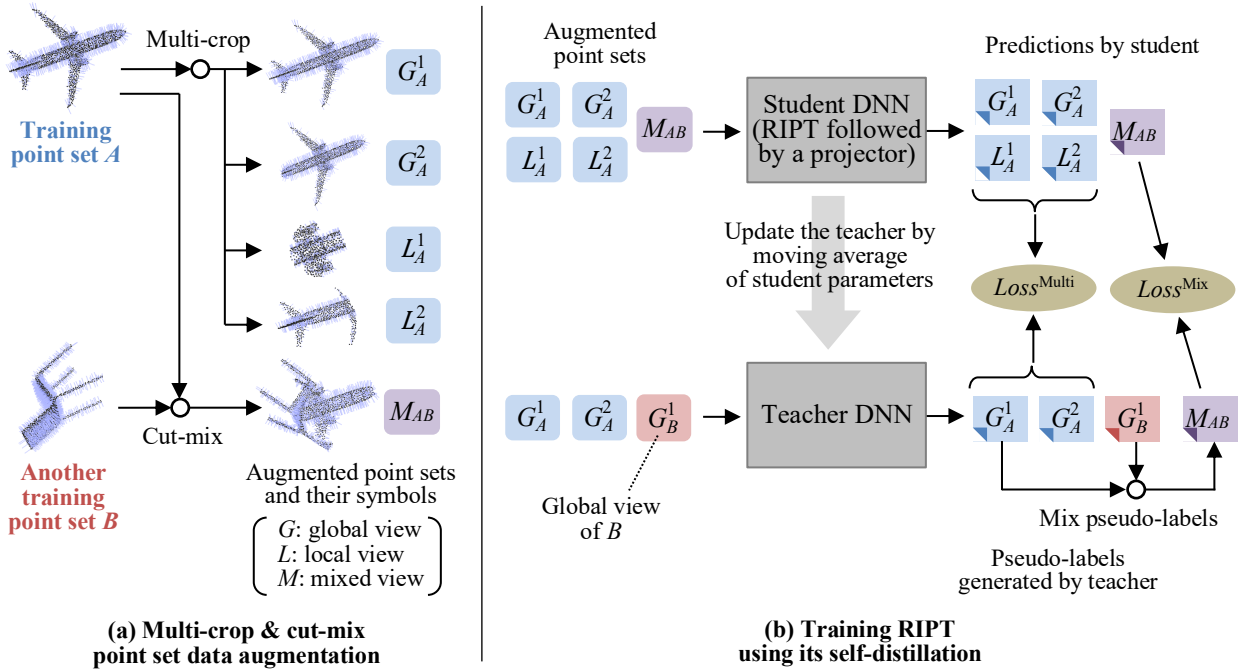

Figure 2. Training framework of the proposed SSL algorithm, i.e., SDMM. (a) Our data augmentation generates diverse 3D point sets including two global views, two local views and one mixed view per training 3D point set. (b) RIPT in the student DNN is trained via the task of pseudo-label classification. Pseudo-labels are generated by using the teacher DNN, whose parameters are computed as the moving average of the student parameters.

## 2. Related work

*2.1. Rotation-invariant analysis of 3D point sets using deep learning*

PointNet [19] is a pioneering end-to-end DNN for 3D point set analysis. Following the success of PointNet, various DNN architectures that incorporate effective shape feature extraction module, such as graph convolution [20] and self-attention [21] have been proposed. However, these DNNs do not have rotation invariance since they directly process raw coordinates of a 3D point set. That is, the accuracy of these DNNs significantly drops when they analyze 3D point sets whose orientations are different from those seen during training. As demonstrated in several studies [6][9][28], the abovementioned DNNs have difficulty in achieving rotation invariance even when they are trained with data augmentation that randomly rotates 3D point sets.

Recently, several DNNs for RI analysis of 3D point sets have been proposed [1]. These DNNs are trained under the supervised learning framework and achieved high accuracy in classification or semantic segmentation of rotated 3D point sets. There are roughly three approaches to achieve rotation invariance. They are, (1) normalizing rotation of global and/or local 3D shapes, (2) extracting inherently RI 3D geometric features, and (3) designing a DNN architecture having rotation equivariance.

**Normalizing rotation of 3D shapes:** Xiao et al. [3] and Li et al. [4] normalize the orientation of an entire 3D point set by using Principal Component Analysis (PCA). That is, three mutually orthogonal axes for rotation normalization are obtained by eigendecomposition of the covariance matrix of the 3D points. The rotation-normalized global 3D point sets are fed into the existing DNN [19][20][21] to achieve RI analysis. However, misalignment of orientations leads to low accuracy. Aligning global 3D point sets to their canonical pose is difficult since they have highly diverse 3D shapes and may include background 3D points sampled at, e.g., floors and walls.

The group of previous work [2][5][22][23] employs rotation normalization of local regions of the 3D point set. They establish a local coordinate system called Local Reference Frame (LRF) to rotation-normalize each local region. The LRF of [2] and [22] is computed by applying PCA to the 3D points within a local region. The LRF of [23] is constructed from the barycenter of the local region and the origin of the global coordinate. [5] employs a DNN to infer the LRF for each local region. The set of rotation-normalized local regions is fed into the DNN for RI 3D point set analysis.

Compared to global 3D point sets whose shape can be complex, orientations of local 3D point sets are easier to normalize as their shapes tend to be simpler. However, as discussed in Section 1, rotating small regions to different orientations causes the pose information loss problem [8]. DLAN [2] mitigates this issue by randomizing the region size to generate regions having medium to large size. DLAN also describes each region by a handcrafted feature having 3D grid structure to preserve the spatial layout of 3D points within the region. Inspired by DLAN, our RIPT samples global-scale regions and describes them by the 3D grid-structured features to alleviate the pose information loss.

**Extracting inherently RI features:** The group of previous studies [6][7][8][9][24][25][26] adopts 3D geometric features that are not influenced by rotation of input 3D point sets. While these studies propose diverse DNN architectures, their basic design principles are similar. That is, the first layer of these DNNs sample local regions, each of which consists of a key 3D point and its neighboring 3D points. Each local region is encoded by the RI low-level features such as distances among 3D points and angles among surface normals. The set of RI local features is then passed to subsequent layers to produce object-level RI global feature. Although being inherently RI, the distance/angle-based features at a local scale cause pose information loss since these features no longer contain such information. In addition, using the low-level features suffers from a significant loss of 3D shape information. Our RIPT, on the other hand, adopts a richer geometric feature having 3D grid structure as input to the DNN.

**Designing rotation-equivariant DNN:** Shen et al. [27], Deng et al. [28], and Assaad et al. [29] propose feature transformation layers having rotation equivariance to extract RI 3D point set features. Rotation equivariance here means that a rotation of a 3D point set in the 3D coordinate space induces the same rotation of the feature in the latent feature space. [27], [28] and [29] extend DNN neurons from 1D to 3D so that they can preserve SO(3) rotation of the input 3D point set. [27] and [29] compute the RI 3D shape feature by an inner product of the rotation-equivariant feature and itself to cancel out the rotation. The RI feature of [28] is computed from the norms of the 3D neurons. These rotation-equivariant DNNs achieve rotation invariance without losing pose information. However, as pointed out by [8], the rotation-equivariant DNNs impose strong constraints (e.g., linearity) on their layers to achieve rotation equivariance, sacrificing flexibility of feature transformation.

Compared to the abovementioned RI DNNs for supervised learning, RI DNNs for SSL have not been sufficiently explored. Deng et al. [12] and Marcon et al. [13] propose autoencoders for SSL of RI local 3D point set features. [12] employs inherently RI 3D geometric features while [13] adopts rotation-equivariant signals as input to the autoencoder, respectively. Note that these studies aim to learn part-level RI features suitable for 3D point set registration. As such, they can't perform analysis of entire 3D shapes such as retrieval, clustering, and classification. To the best of our knowledge, there has been no prior work specifically focusing on SSL of object-level RI 3D point set features usable for analyzing entire 3D shapes. Our study aims at acquiring such object-level RI features under the SSL setting.

*2.2. Self-supervised learning of feature representations*

*2.2.1. Self-supervised feature learning for 2D images*

SSL has been actively studied especially in the field of 2D image analysis [30]. A wide variety of pretext tasks including self-reconstruction [31], pseudo-label classification [32], and feature contrast [33] have been developed. Recently, knowledge distillation has been introduced to SSL of visual features to boost their accuracy [34].

**Self-reconstruction:** Autoencoder [31] learns features via the self-reconstruction task. An autoencoder is formed by pairing an encoder DNN with a decoder DNN. The encoder embeds an entire input 2D image into the latent feature space while the decoder reconstructs the input from the latent feature. He et al. [35] propose masked autoencoding suitable for learning contexts among local patches of a 2D image. [35] employs Vision Transformer (ViT) [36], which is built upon a self-attention mechanism, as an encoder DNN.

**Pseudo-label classification:** This approach automatically assigns a pseudo-label to each unlabeled training 2D image. Instance Discrimination [32][37] creates a pseudo-label that is unique per training sample. Pseudo-labels of DeepCluster [38] are generated by applying *k*-means clustering to the set of latent features extracted by the encoder DNN. SwAV [18] and Local Aggregation [39] incorporate the clustering procedure as a part of the encoder DNN to generate pseudo-labels more adapted to the latent features.

**Feature contrast:** This approach, also known as contrastive learning, trains an encoder DNN by comparing latent features extracted by the encoder. The feature contrast methods try to form a latent feature space where features of positive sample pairs are embedded closer while features of negative sample pairs are embedded further apart from each other. Majority of the feature contrast methods, e.g., [33], [40], [41], [42] creates a positive pair by coupling two different "views" of a training 2D image. These two views are generated by applying data augmentation with different augmentation parameters to the training sample. A negative pair is formed between two views derived from mutually different training samples. SimCLR [40] uses large minibatches to effectively learn distance metric among positive/negative pairs. MoCo [41] employs a large dictionary of latent features so that many latent features can be used for their contrast. SimSiam [42] enables training with small minibatches consisting only of positive pairs by introducing a Siamese but asymmetric encoder architecture.

**Knowledge distillation:** Knowledge distillation [43] was originally proposed for model compression of supervisedly trained DNNs. In SSL using knowledge distillation [16][34], the teacher and the student have an identical DNN architecture, but different parameters. During training, the parameters of the teacher are usually computed as a moving average of student parameters over past training steps. Such a "temporal ensemble" of previous students enables the teacher to produce expressive features useful for training of the current student. BYOL [34] incorporates knowledge distillation into the feature contrast framework. BYOL encourages the latent features extracted by the student encoder to be close to those extracted by the teacher encoder. DINO [16] is a distillation-based SSL algorithm similar to BYOL. The pretext task of DINO can be interpreted as pseudo-label classification since an output from the teacher of DINO is represented as a categorical distribution. DINO also employs the MultiCrop data augmentation [18] to create a minibatch having diverse whole and partial 2D images. More recently, Ren et al. [44] incorporated the CutMix data augmentation [17] into DINO to create highly diverse 2D images for training. ViT trained with the method [44] achieves state-of-the-art feature accuracy. This paper extends the idea of [44] to 3D point sets to effectively train our Transformer-based DNN, i.e., RIPT.

*2.2.2. Self-supervised feature learning for 3D point sets*

Various algorithms for SSL of 3D point set features have been proposed [11]. These algorithms usually adopt learning frameworks similar to SSL for 2D images. Several studies [45][46][47][48] employ self-reconstruction as a pretext task. FoldingNet by Yang et al. [45] and Canonical Capsules by Sun et al. [46] are autoencoders for 3D point set data designed for learning 3D point set features. Point-BERT [47] and MaskPoint [48] utilize the masked autoencoding framework to train a Transformer-based DNN for 3D point sets. Wang et al. [49] and Fu et al. [50] adopt pseudo-label classification. [49] applies Instance Discrimination to 3D point sets. DCGLR [50] incorporates the idea of DINO to SSL of 3D point set features. Another group of studies [51][52][53] adopts the feature contrast approach. [51] and [52] extend SimCLR to SSL of 3D point set features, while [53] utilizes diffusion distance to contrast features extracted by an encoder DNN.

The abovementioned SSL algorithms for 3D point set analysis have succeeded in learning point set features useful for retrieval, classification, segmentation, of 3D objects. However, these features are not invariant to SO(3) rotations of 3D objects. Our approach is novel in that it is capable of learning RI object-level point set features under the SSL framework. Note that our SDMM is similar to DCGLR by Fu et al. [50] since both algorithms are built upon the idea of DINO. Yet, our SDMM takes a step further. That is, by incorporating the idea of CutMix data augmentation [17], our SDMM can generate more diverse training data to acquire accurate 3D point set features.

## 3. Proposed algorithm

*3.1. Overview of the proposed algorithm*

To obtain RI and accurate object-level 3D point set features without relying on semantic labels, we propose a DNN architecture called RIPT and an SSL algorithm called SDMM. Figure 1 illustrates the architecture of RIPT. Figure 2 depicts the training framework of SDMM.

We design RIPT to satisfy the three criteria: rotation invariance, high accuracy, and high efficiency. To achieve rotation invariance, we adopt an approach similar to the existing RI DNNs. Specifically, RIPT samples multiple regions from an input 3D point set and normalizes the rotation of these regions based on their local 3D geometry, i.e., normal orientation of the region. To obtain accurate features, we address the problem of pose information loss by sampling global regions instead of local ones. Each rotation-normalized global region is described by the handcrafted geometric feature [2][54] having a 3D grid structure to preserve the spatial layout of partial shapes within the global region. The 3D grid features are then transformed into feature vectors, or tokens, by a fully-connected (FC) layer. The set of tokens is refined and then aggregated to an object-level RI latent feature by using the vector self-attention mechanism [15] that can consider the relationships across tokens and among channels of tokens. The commonly used scalar self-attention [14], in contrast, only considers relations among tokens. To achieve high efficiency, i.e., high computation speed and low memory footprint, RIPT is composed of a small number of feature transformation layers. RIPT has only one FC layer and two self-attention blocks for token feature extraction and token set refinement, respectively. In addition, each self-attention block downsamples the tokens to reduce the computation cost for their refinement.

To effectively train RIPT, SDMM leverages diverse training 3D point sets created by the multi-crop and cut-mix data augmentation. As shown in Figure 2a, our data augmentation creates three types of augmented 3D point sets, or "views", for each training 3D point set. They are global view, local view, and mixed view. Our multi-crop cuts out a (nearly) whole 3D shape as a global view and a partial 3D shape as a local view. Our cut-mix generates a mixed view by taking a union of two subsets sampled from two different training 3D point sets [55]. These three views are used as input data for the self-distillation framework. As shown in Figure 2b, we use the student DNN and teacher DNN having the identical architecture, i.e., RIPT followed by a projector. The role of the teacher DNN is to generate pseudo-labels, which are used as supervision signals for the student DNN. The outputs from the teacher are directly used as the pseudo-labels for the global and local views. The pseudo-label for the mixed view is created by mixing the two pseudo-labels derived from mutually different training point sets. The parameters of the student DNN are tuned so that the predictions by the student match their corresponding pseudo-labels generated by the teacher. At every training step, the parameters of the teacher DNN are updated as the moving average of the student parameters. After the training, we employ RIPT in the teacher DNN (not in the student DNN) as a feature extractor since the teacher is a temporal ensemble of the students over training iterations. Such ensemble model is expected to converge to a better solution and exhibit a higher generalization ability compared to the non-ensemble model, i.e., the student.

*3.2. Rotation-Invariant Point set token Transformer (RIPT)*

RIPT takes as its input an oriented 3D point set denoted as $P = \{(\mathbf{p}_i, \mathbf{o}_i)\}_{i=1, ..., N}$. $P$ consists of $N$ (e.g., $N = 1,024$) oriented 3D points, where the coordinates $\mathbf{p}_i$ of the $i$-th 3D point is associated with its 3D orientation vector $\mathbf{o}_i$. When $P$ is a realistic 3D point set created by using a scanner device, the orientation $\mathbf{o}_i$ can be a unit vector perpendicular to the tangent plane estimated from the local geometry around $\mathbf{p}_i$. When $P$ is a synthetic 3D point set created from a polygonal 3D CAD model, $\mathbf{o}_i$ can be the normal vector of the polygon on which $\mathbf{p}_i$ is sampled. An RI latent feature of $P$ is extracted by feeding $P$ to RIPT, which consists of two sub-networks, i.e., Rotation-Invariant Tokenizer (RI-Tokenizer) followed by Token Set Transformer (TS-Transformer).

*3.2.1. RI-Tokenizer*

RI-Tokenizer transforms $P$ into a token set which consists of $T$ (e.g., $T = 256$) token points and their corresponding RI token features. First, we use Farthest Point Sampling (FPS) [56] to choose $T$ token points from the $N$ 3D points in $P$. For each token point $\mathbf{c}$, we sample a global region that contains all the $N$ oriented points. We then normalize the rotation of the global region by using the LRF centered at $\mathbf{c}$. Each LRF is comprised of three mutually orthogonal unit vectors, i.e., $\mathbf{u}_1$, $\mathbf{u}_2$, and $\mathbf{u}_3$. We use the orientation vector of $\mathbf{c}$ as $\mathbf{u}_1$. $\mathbf{u}_2$ is computed by using PCA of the 3D points within the global region. We construct the covariance matrix $\mathbf{M}$ that captures the local geometry around the token point $\mathbf{c}$ in 3D (In [57], $\mathbf{c}$ is called feature point). Specifically, $\mathbf{M}$ is computed as:

$$\mathbf{M} = \frac{1}{\sum_{i=1}^{N}(r-d_i)}\sum_{i=1}^{N}(r-d_i)(\mathbf{p}_i-\mathbf{c})(\mathbf{p}_i-\mathbf{c})^T \qquad (1)$$

where $d_i$ is the Euclidean distance between $\mathbf{p}_i$ and $\mathbf{c}$, and $r$ is the maximum of $d_i$ values computed for the $N$ points. We apply eigen decomposition to $\mathbf{M}$ and pick up the eigen vector associated with the second largest eigen value. $\mathbf{u}_2$ is obtained by projecting the eigen vector onto the plane perpendicular to $\mathbf{u}_1$. $\mathbf{u}_3$ is computed as a cross product of $\mathbf{u}_1$ and $\mathbf{u}_2$. The LRF is represented as a 3×3 matrix $\mathbf{R}$, which is created by stacking $\mathbf{u}_1$, $\mathbf{u}_2$, and $\mathbf{u}_3$. $\mathbf{R}$ is used to normalize rotation of the global region. The oriented 3D points in the global region after rotation normalization is then denoted as $\{((\mathbf{p}_i - \mathbf{c})\cdot\mathbf{R}, \mathbf{o}_i\cdot\mathbf{R})\}_{i=1, ..., N}$.

Each rotation-normalized region is described by using the handcrafted 3D geometric feature called Position and Orientation Distribution (POD) [54]. Specifically, the oriented 3D points in the region are spatially partitioned by using regular 3D grids. This paper uses 6×6×6 grids, which is finer than 4×2×1 grids of the original POD, to preserve the spatial layout of the 3D points scattered in the global region. For each cell in the 3D grids, we compute 10D features, that are, frequency of points (1D), mean of point coordinates (3D), and covariance of orientation vectors (6D), by using the oriented 3D points that lie within the cell. The 3D grid-structured RI feature is then converted into a token feature by a simple operation similar to ViT [36]. That is, the 3D grid feature is flattened to a vector having 6×6×6×10 = 2,160D, and then linearly projected by using a single FC layer to produce the token feature $\mathbf{x}$ having 512D. The token feature $\mathbf{x}$ is paired with its corresponding token point $\mathbf{c}$. Collecting all the $T$ pairs of $\mathbf{c}$ and $\mathbf{x}$ forms the token set denoted by $S = \{(\mathbf{c}_i, \mathbf{x}_i)\}_{i=1, ..., T}$ for the input $P$.

### 3.2.2. TS-Transformer

TS-Transformer refines the token features contained in the token set $S$ and then aggregates them to a single RI feature vector for $P$, which is the input oriented 3D point set. TS-Transformer uses two self-attention (SA) blocks for token feature refinement. Each SA block employs vector self-attention [15]. Vector self-attention is capable of highly flexible feature transformation since its attention map is computed adaptively to the input token features, considering the relationships both between the tokens and between the feature channels. Yet, applying vector self-attention to all the $T$ tokens incurs high temporal and spatial complexities since these costs increase quadratically with $T$, i.e., $O(T^2)$. We thus reduce the computation costs by applying vector self-attention to only $k$ nearest neighbors of each token point. Such "localized" self-attention can reduce both temporal and spatial costs down to $O(k \cdot T)$. We use $k = 4$ for the first SA block and $k = 8$ for the second block to gradually increase the receptive field size.

Figure 3 illustrates the processing pipeline of each SA block. Let the input token set having $T_{in}$ tokens be $S_{in} = \{(\mathbf{c}_i, \mathbf{x}_i)\}_{i=1,\ldots,T_{in}}$. We first subsample $S_{in}$ by applying FPS to the $T_{in}$ 3D token points to create the subset of $S_{in}$ denoted by $S'_{in} = \{(\mathbf{c}'_i, \mathbf{x}'_i)\}_{i=1,\ldots,T_{out}}$. We set $T_{out} = T_{in} / 2$ to reduce the computation cost of the SA block. We then utilize vector self-attention to refine the token features. Specifically, the refined token feature $\mathbf{y}'_i$ for the token $(\mathbf{c}'_i, \mathbf{x}'_i)$ in $S'_{in}$ is computed as:

$$\mathbf{y}'_i = \sum_{\mathbf{x}_j \in knnFeats(\mathbf{c}'_i)} \rho\left(\alpha(\mathbf{x}'_i) - \beta(\mathbf{x}_j)\right) \odot \gamma(\mathbf{x}_j) \quad (2)$$

where the *knnFeats* function finds, from $S_{in}$, a set of $k$ token features whose token points are $k$ nearest neighbors of $\mathbf{c}'_i$. Each of $\alpha$, $\beta$, and $\gamma$ is a linear projection function implemented as an FC layer. $\rho$ is the softmax function, whose output is the attention map. $\odot$ denotes elementwise multiplication. We establish a residual connection [58] between the input and output of Eq. 2, i.e., $\mathbf{y}'_i + \mathbf{x}'_i$, to stabilize DNN training. The result of addition is further processed by a Batch Normalization (BN) [59] layer and its subsequent two FC layers with ReLU activation to produce the output token feature $\mathbf{f}'_i$. $\mathbf{x}'_i$ in $S'_{in}$ is replaced with $\mathbf{f}'_i$ to create the output token set $S_{out} = \{(\mathbf{c}'_i, \mathbf{f}'_i)\}_{i=1,\ldots,T_{out}}$. $S_{out}$ is used as the input to the next SA block or its subsequent feature aggregation. We fix the number of neurons for all the FC layers in the SA block at 512.

The token set $S_{out}$ refined by the second SA block is aggregated to produce an object-level feature per 3D point set. When RI-Tokenizer samples $T = 256$ tokens, the token set $S_{out}$ produced by the second SA block consists of 64 tokens. We discard the 3D token points $\{\mathbf{c}'_i\}_{i=1,\ldots,64}$ in $S_{out}$ and aggregate the 512D token features $\{\mathbf{f}'_i\}_{i=1,\ldots,64}$ by using average pooling. The aggregated 512D feature is compressed by using an FC layer down to 256D. The compressed feature is then normalized by its L2 norm to produce the RI latent feature for the input $P$. In the training phase, the RI latent features are fed into the projector DNN to generate pseudo-labels. In the testing phase, the RI latent features are used as 3D point set features for their comparison.

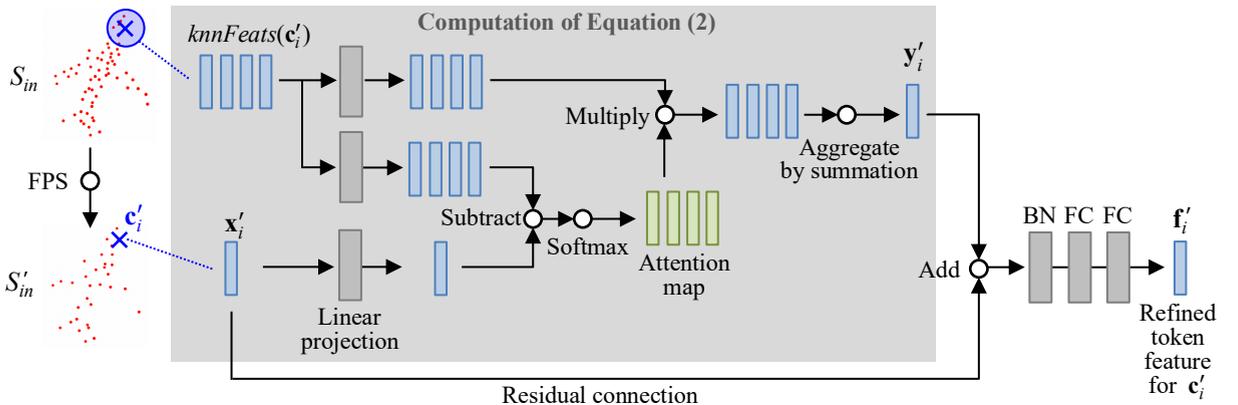

Figure 3. Processing pipeline of our self-attention (SA) block in TS-Transformer. We use the localized vector self-attention to effectively and efficiently refine the set of RI token features. Each SA block subsamples the token points by using Farthest Point Sampling (FPS) to reduce overall computation cost of TS-Transformer.

*3.3. Self-Distillation with Multi-crop and cut-Mix point set augmentation (SDMM)*

*3.3.1. Multi-crop and cut-mix point set augmentation*

For each training 3D point set $A$, SDMM creates two global views ($G_A^1$ and $G_A^2$), two local views ($L_A^1$ and $L_A^2$), and one mixed view ($M_{AB}$). We assume that $A$ consists of $n = 1,024$ oriented 3D points. We first normalize the position and scale of $A$. Specifically, $A$ is translated so that its gravity center coincides the origin of the 3D space, and then is uniformly scaled so that its entirety is enclosed by a unit sphere.

Each global/local view of $A$ is created by using the following cropping procedure. We randomly choose one of the 3D points in $A$. We then sample the $\lfloor rn \rfloor$ neighbors of the chosen 3D point to create a global/local view of $A$. $r$ for the global view is randomly sampled from the uniform distribution $U(0.6, 1)$. $r$ for the local view is randomly sampled from $U(0.4, 0.6)$. After cropping, we randomly subsample or duplicate the 3D points so that each global view consists of 1,024 3D points and each local view contains 512 3D points.

The mixed view for $A$ is created as follows. We first pick up another training 3D point set $B$ from the training minibatch. Since we assume that the orientations of the training 3D point sets are inconsistent, we randomly rotate $B$ even if its original orientation is aligned. The mixed view $M_{AB}$ is created by taking the union of the two subsets, i.e., $A'$ and $B'$, sampled from $A$ and $B$ respectively:

$$M_{AB} = A' \cup B'$$
$$A' = \{ \mathbf{p}_i \mid (\mathbf{p}_i \text{ is one of the } \lfloor mn \rfloor \text{ nearest points of } \mathbf{p}) \wedge (\mathbf{p}_i \in A) \}_{i=1, \ldots, \lfloor mn \rfloor} \quad (3)$$
$$B' = \{ \mathbf{p}_i \mid (\mathbf{p}_i \text{ is one of the } n - \lfloor mn \rfloor \text{ farthest points of } \mathbf{p}) \wedge (\mathbf{p}_i \in B) \}_{i=1, \ldots, n-\lfloor mn \rfloor}$$

In Equation 3, $n$ is the number of 3D points per point set, i.e., $n = 1,024$. $m$ is the mixing parameter which is randomly sampled from $U(0, 1)$. The 3D point $\mathbf{p}$, which works as a reference point for cut-mix, is randomly chosen from $A$. Each mixed view consists of $n = 1,024$ 3D points.

To further diversify the 3D shapes of the five views, i.e., $G_A^1$, $G_A^2$, $L_A^1$, $L_A^2$, and $M_{AB}$, the 3D point set of each view is randomly and anisotropically scaled. The mutually orthogonal three axes for anisotropic scaling are chosen randomly per view. The scaling factor for each axis is randomly sampled from $U(0.67, 1.5)$. Each anisotropically scaled view is treated as $P$ defined in Section 3.2 and is fed into RIPT to extract the RI latent feature.

**Differences from existing 3D point set mixup:** The existing mixup algorithms for 3D point sets [55][72] first find point-to-point correspondence between two different 3D point sets. PointMixup [72] interpolates between the corresponding two points to generate a new point set. To create a mixed pointset from a pair of point sets, PointCutMix [55] chooses points from either one of the two point sets. When two input 3D point sets belong to the same semantic category and their orientation are aligned, such correspondence-based mixup algorithms can generate meaningful 3D shapes preserving the same semantics. However, in our case, establishing point-to-point correspondences would be less effective since our cut-mix method combines two point sets having different orientations even if they belong to the same category. In addition, computing point-to-point correspondences requires high temporal cost and may become a temporal bottleneck during training. We thus employ the simple cut-mix strategy shown in Equation 3, which requires nearest/farthest point search only. Furthermore, labels to be mixed in our unsupervised setting differ from those for a supervised setting. That is, [55] and [72] mix semantic labels annotated by humans since they are intended for supervised learning. In contrast, ours mixes pseudo-labels generated by the teacher DNN for the effective SSL, which will be described in the next subsection.

*3.3.2. Self-supervised feature learning using self-distillation*

We use the student DNN $f_s$ parameterized by $\theta_s$ and the teacher DNN $f_t$ parameterized by $\theta_t$. The parameters of the student, i.e., $\theta_s$, are the optimization target of the training by SDMM. The parameters of the teacher, i.e., $\theta_t$, are updated, at every training step, by calculating $\theta_t \leftarrow \lambda \theta_t + (1 - \lambda)\theta_s$. During training, the value for $\lambda$ is increased from 0.996 to 1 by using a cosine scheduling [16]. Both the student and teacher DNNs are composed of RIPT followed by a projector DNN. The projector DNN consists of three FC layers having 1,024, 128, $H$ neurons, respectively. $H$ corresponds to the dimensionality of pseudo-labels, and is set at 1,024 in this paper. The neurons at the first and second FC layers are activated by the GELU function [60].

Given the input oriented point set $P$, the teacher outputs the pseudo-label of $P$, while the student outputs the prediction for $P$. Each output from the student or the teacher is represented as a categorical distribution $\mathbf{d}$ having $H$ dimensions. The prediction by the student for the input $P$, i.e., $\mathbf{d}_s(P)$, is computed by the following equation:

$$\mathbf{d}_s(P)^{(i)} = \frac{exp(f_s(P)^{(i)}/\tau_s)}{\sum_{h=1}^{H} exp(f_s(P)^{(h)}/\tau_s)} \quad (4)$$

where $\cdot^{(i)}$ denotes the *i*-th element of a vector and *exp* is the exponential function. $\tau_s$ is the temperature parameter which is fixed at 0.4 in this paper. The pseudo-label $\mathbf{d}_t(P)$ generated by the teacher is computed as follows.

$$\mathbf{d}_t(P)^{(i)} = \frac{exp((f_t(P)^{(i)} - \boldsymbol{\mu}^{(i)})/\tau_t)}{\sum_{h=1}^{H} exp((f_t(P)^{(h)} - \boldsymbol{\mu}^{(h)})/\tau_t)} \quad (5)$$

In Equation 5, subtraction by $\boldsymbol{\mu}$ translates the outputs of the teacher to around the origin of the $H$ dimensional space. This translation operation prevents the activations of $f_t(P)$ from being biased towards specific dimensions [16]. $\boldsymbol{\mu}$ is initialized by a zero vector and is updated at every training step by $\boldsymbol{\mu} \leftarrow 0.9\boldsymbol{\mu} + 0.1\boldsymbol{\mu}_{mb}$, where $\boldsymbol{\mu}_{mb}$ is the mean of teacher outputs computed from an input minibatch. $\tau_t$ is the temperature parameter for the teacher. We use $\tau_t = 0.1$ to generate categorical distributions sharper than those predicted by the student.

To construct a training minibatch, we first randomly sample $E$ (we use $E = 32$) oriented 3D point sets from the training dataset. Each training sample $A$ is augmented by using the method described in Section 3.3.1 to create five views (i.e., $G_A^1$, $G_A^2$, $L_A^1$, $L_A^2$, and $M_{AB}$). Therefore, each training minibatch consists of $5E$ oriented point sets. The teacher DNN processes the global views to generate the pseudo-labels. These pseudo-labels are used as supervision signals for the global views and the local views. The pseudo-label for each mixed view is computed by linearly combining the pseudo-labels for the two global views $G_A^1$ and $G_B^1$:

$$\mathbf{d}_{AB} = m\mathbf{d}_t(G_A^1) + (1 - m)\mathbf{d}_t(G_B^1) \quad (6)$$

In Equation 6, $m$ is the mixing parameter used in the computation of Equation 3. The loss value for the training sample $A$ is computed as $Loss^{Multi} + Loss^{Mix}$ where:

$$Loss^{Multi} = \sum_{P \in \{G_A^1, G_A^2\}} \sum_{\substack{P' \in \{G_A^1, G_A^2, L_A^1, L_A^2\} \\ \text{and } P' \neq P}} l\left(\mathbf{d}_t(P), \mathbf{d}_s(P')\right) \quad (7)$$

$$Loss^{Mix} = l\left(\mathbf{d}_{AB}, \mathbf{d}_s(M_{AB})\right) \quad (8)$$

The function $l$ in Equation 7 and Equation 8 is a cross-entropy between two categorical distributions. The loss values computed for all the $E$ training samples in the minibatch are averaged to obtain an overall loss value. The student parameters $\theta_s$ are updated to decrease the overall loss. For optimization, we use Adam [61] with warmup [62]. Specifically, the learning rate is initialized at 0.0001 and is linearly increased to 0.0005 during the first 20 epochs. After the 20-th epoch, the learning rate is decreased to 0.0001 by using a cosine scheduling. The total number of training epochs is set to 200.

After the training, RIPT in the teacher DNN is used as a feature extractor. Given a testing oriented 3D point set, we normalize its position and scale by using the method described in Section 3.3.1. Note that the multi-crop and cut-mix augmentations are not applied to testing 3D point sets. The normalized testing 3D point sets are fed into the trained RIPT to extract their RI latent features.

## 4. Experiments and results

### 4.1. Experimental setup

We quantitatively evaluate rotation invariance and accuracy of the 3D point features learned by our algorithm, i.e., RIPT and SDMM. We also evaluate training efficiency of the RIPT architecture via comparison against the existing DNN architectures having rotation invariance.

**Benchmark datasets:** Most of our experiments are conducted by using the four benchmark datasets: ModelNet10 (MN10) [63], ModelNet40 (MN40) [63], ScanObjectNN OBJ_ONLY (SO-OO) [64], and ScanObjectNN OBJ_BG (SO-OB) [64]. In addition, some experiments use two larger datasets, i.e., ShapeNetCore55 (SN55) [65] and ScanObjectNN PB_T50_RS (SO-PB) [64]. Table 1 summarizes the statistics of the datasets. For all six datasets, each data sample is represented as an oriented 3D point set $P$ consisting of $N = 1,024$ 3D points, where each point **p** has an orientation vector **o**. Each sample of MN10, MN40, and SN55 is a synthetic 3D point set created from a polygonal 3D CAD model. To convert a polygonal 3D model to an oriented 3D point set, we randomly and uniformly sample a set of $N = 1,024$ 3D points from the surface of the polygonal model by using the algorithm by Ohbuchi et al. [66]. The orientation vector, i.e., **o**, for each 3D point **p** is computed as the normal vector of the polygon on which **p** is sampled. On the other hand, each data sample in SO-OO, SO-OB, and SO-PB is a realistic 3D point set created by scanning a real-world 3D object. The ScanObjectNN dataset provides a set of 2,048 3D points without orientations per 3D object.

In our experiments, we estimate the orientation vector for each 3D point **p** by using PCA. Specifically, PCA is applied to every local region consisting of **p** and its 16 neighbors. The eigen vector associated with the smallest eigen value becomes the orientation vector **o** for **p**. The sign of the eigen vector is disambiguated by using the algorithm by Salti et al. [57]. We then randomly subsample $N = 1,024$ 3D points from the 2,048 3D points per 3D object. Note that the semantic labels attached to the training samples are not used for training since we aim at SSL of 3D point set features.

**Rotation settings:** To evaluate rotation invariance of the learned 3D point set features, we use the three rotation settings, that are, "Nr/Nr", "Nr/Rr", and "Rr/Rr". Nr, which stands for No rotation, indicates that a DNN consumes 3D point sets whose poses are consistently aligned. Rr, short for Random rotation, means that each 3D point set is randomly rotated by the rotation matrix sampled from SO(3). Nr/Nr (or Rr/Rr) denotes that no rotations (or random rotations) are applied to both training and testing 3D point sets. In the Nr/Rr setting, training 3D point sets are not rotated, while testing 3D point sets are randomly rotated. 3D point set features having rotation invariance should produce nearly equal accuracy values for the three rotation settings.

**Accuracy measure:** In the majority of the experiments, we evaluate accuracy of the learned 3D point set features under the information retrieval scenario. Retrieval accuracy is evaluated by using macro Mean Average Precision (macroMAP). A macroMAP is computed by the following procedure. Each 3D point set in the testing dataset is chosen as a retrieval query and remaining 3D point sets in the testing dataset are treated as retrieval targets. We use the Euclidean distance to compute distances among 3D point set features for ranked retrieval. We compute the Average Precision (AP) for each query by using the semantic labels of the testing 3D point sets as groundtruths. The AP values are averaged for each semantic category to produce category-wise AP values. These category-wise AP values are averaged over categories to obtain the macroMAP for the testing dataset.

To evaluate the generalizability of the learned object-level features, we also conduct experiments using classification and clustering scenarios. For the classification scenario, a linear SVM [69] is trained on the 3D point set features extracted from the training samples. We report macro-averaged classification accuracy for the testing samples. A macro-averaged classification accuracy is obtained by calculating the classification accuracy for each semantic category and then averaging these accuracy values across categories. For the clustering scenario, $k$-means algorithm is applied to the features of the testing samples where $k$ corresponds to the number of semantic categories listed in Table 1. Clustering accuracy is measured by Normalized Mutual Information (NMI), which quantifies the similarity between clusters generated by $k$-means and groundtruth clusters. NMI ranges from 0 to 1, with a higher value indicating better clustering accuracy.

We implemented RIPT and SDMM by using Python and PyTorch library [67]. Most of the experiments were done on a PC having a CPU (*AMD Ryzen 9 5900X*) and two GPUs (*Nvidia GeForce RTX 3090* and *RTX TITAN*). Each GPU has 24 GBytes of VRAM. Our proposed algorithm runs on a single GPU, while some experiments using an existing DNN require two GPUs to store DNN parameters and feature maps on VRAM.

Table 1. Statistics of the datasets used in our experiments.

| Dataset | ModelNet10 (MN10) | ModelNet40 (MN40) | ShapeNetCore55 (SN55) | ScanObjectNN OBJ_ONLY (SO-OO) | ScanObjectNN OBJ_BG (SO-OB) | ScanObjectNN PB_T50_RS (SO-PB) |
|---|---|---|---|---|---|---|
| Type of 3D point sets | Synthetic | Synthetic | Synthetic | Realistic | Realistic | Realistic |
| # of semantic categories | 10 | 40 | 55 | 15 | 15 | 15 |
| # of training samples | 3,991 | 9,840 | 35,764 | 2,309 | 2,309 | 11,416 |
| # of testing samples | 908 | 2,468 | 10,265 | 581 | 581 | 2,882 |

## 4.2. Experimental results and discussions

### 4.2.1. Comparison against existing algorithms

**Comparison against existing SSL algorithms for 3D point sets:** We evaluate rotation invariance and accuracy of the features learned by our proposed algorithm, i.e., RIPT + SDMM. As competitors, we chose six SSL algorithms for 3D point sets, that are, FoldingNet [45], Canonical Capsules [46], Point-BERT [47], MaskPoint [48], DCGLR [50], and PPF-FoldNet [12]. Note that the algorithms except for PPF-FoldNet use a DNN architecture without rotation invariance. Since PPF-FoldNet was originally proposed to learn part-level RI features, we modify PPF-FoldNet to learn object-level RI features. Specifically, we enlarge the region size for extracting inherently RI geometric features from local to global, and train the DNN by reconstructing the global RI geometric features.

Table 2 compares retrieval accuracies of the features obtained by the self-supervised 3D point set feature learning algorithms. For all the four datasets, the SSL algorithms using a non-RI DNN architecture show a large gap in retrieval accuracies between the Nr/Nr setting and the other two rotation settings. Since these non-RI DNNs extract features heavily depending on the pose of an input 3D point set, their retrieval accuracies suffer when testing 3D point sets are randomly rotated. On the other hand, PPF-FoldNet and RIPT + SDMM exhibit rotation invariance since they produce nearly equal retrieval accuracies for the three rotation settings. The small variance in accuracy between the three rotation settings is due to the randomness during training, such as initialization of DNN parameters and creation of minibatches. Our RIPT + SDMM outperforms PPF-FoldNet by a large margin. The superiority of RIPT + SDMM demonstrates the soundness of our design principle for learning object-level RI point set features. For the SO-OO and SO-OB datasets including realistic 3D point sets, our algorithm again yields accuracies significantly higher than the existing algorithms, regardless of their RI properties. Each realistic 3D point set contains 3D points derived not only from a 3D object, but also from its background and measurement noise. We suspect that the existing DNNs which directly process 3D point coordinates are negatively affected by the background and noise, resulting in low accuracy even under the Nr/Nr setting. In contrast, our RIPT could reduce the influence of background and noise by the two processing steps. First, prior to input to the DNN, the 3D grid feature used in RI-Tokenizer can merge background/noise 3D points with their neighboring 3D points sampled from a 3D object. Such weakly encoded geometric features may suppress the influence of background/noise 3D points. Second, vector self-attention in TS-Transformer may inhibit neuron activations corresponding to features extracted from background/noise 3D points.

Table 2. Retrieval accuracy (macroMAP [%]) of features obtained by SSL algorithms for 3D point sets.

| SSL algorithms | MN10 dataset | | | MN40 dataset | | | SO-OO dataset | | | SO-OB dataset | | |
|---|---|---|---|---|---|---|---|---|---|---|---|---|
| | Nr/Nr | Nr/Rr | Rr/Rr | Nr/Nr | Nr/Rr | Rr/Rr | Nr/Nr | Nr/Rr | Rr/Rr | Nr/Nr | Nr/Rr | Rr/Rr |
| FoldingNet [45] | 67.4 | 18.1 | 20.0 | 39.3 | 5.8 | 6.7 | 20.9 | 9.8 | 9.8 | 19.8 | 9.9 | 9.7 |
| Canonical Capsules [46] | 61.2 | 21.0 | 21.7 | 33.9 | 11.2 | 12.1 | 19.6 | 10.3 | 10.1 | 17.9 | 9.2 | 9.2 |
| Point-BERT [47] | 45.6 | 21.3 | 21.2 | 35.1 | 8.5 | 9.0 | 24.3 | 11.0 | 11.3 | 19.4 | 10.1 | 10.6 |
| MaskPoint [48] | 40.5 | 17.9 | 18.6 | 39.8 | 6.1 | 12.5 | 29.8 | 11.0 | 10.4 | 19.8 | 9.4 | 11.2 |
| DCGLR [50] | **76.7** | 19.6 | 27.6 | **53.1** | 9.4 | 11.8 | 28.1 | 10.8 | 11.0 | 25.4 | 12.1 | 12.0 |
| PPF-FoldNet [12] | 36.8 | 37.0 | 37.3 | 23.4 | 23.4 | 23.5 | 18.2 | 18.4 | 18.4 | 14.6 | 15.1 | 15.2 |
| RIPT + SDMM (ours) | 70.3 | **70.5** | **70.4** | 52.3 | **52.0** | **52.1** | **49.7** | **49.8** | **49.6** | **47.3** | **47.3** | **47.5** |

**Comparison against existing DNN architectures having rotation invariance:** To verify the effectiveness of the architecture of RIPT, we compare RIPT with the existing RI DNN architectures designed for supervised learning. For a fair comparison, we vary only a DNN architecture while fixing other experimental settings. That is, we use the same SSL algorithm, i.e., SDMM, across all the DNN architectures. The minibatch size, data augmentation method and learning rate schedule are the same as those described in Section 3.3. Note that the SDMM in this experiment does not use mixed views in its input minibatches. This is because some RI DNNs raise an out-of-GPU-memory error when they tries to process a minibatch containing all the global, local, and mixed views. (This omission of mixed views made accuracies of RIPT in Table 3 lower than those in Table 2).

Table 3. Retrieval accuracy (macroMAP [%]) of features learned by different RI DNN architectures.

| Group | DNN architectures | MN10 dataset | | | MN40 dataset | | | SO-OO dataset | | | SO-OB dataset | | |
|---|---|---|---|---|---|---|---|---|---|---|---|---|---|
| | | Nr/Nr | Nr/Rr | Rr/Rr | Nr/Nr | Nr/Rr | Rr/Rr | Nr/Nr | Nr/Rr | Rr/Rr | Nr/Nr | Nr/Rr | Rr/Rr |
| A | LGR-Net [24] | 28.2 | 28.9 | 28.9 | 20.5 | 20.1 | 20.4 | 16.3 | 16.0 | 16.2 | 17.2 | 16.9 | 17.2 |
| | RIConv++ [9] | 36.8 | 37.5 | 37.3 | 26.6 | 26.7 | 25.2 | 22.2 | 23.4 | 23.2 | 24.3 | 24.9 | 23.1 |
| | PaRI-Conv [8] | 18.3 | 17.7 | 18.6 | 12.0 | 12.3 | 12.3 | 15.8 | 14.8 | 15.9 | 14.9 | 14.8 | 14.8 |
| B | VN-PointNet [28] | 29.6 | 30.7 | 30.7 | 23.1 | 22.3 | 23.2 | 13.8 | 14.6 | 14.4 | 13.7 | 14.2 | 14.2 |
| | VN-DGCNN [28] | 28.7 | 28.3 | 29.0 | 22.4 | 21.2 | 21.4 | 15.6 | 16.1 | 16.0 | 14.5 | 14.3 | 14.5 |
| | REQNN [27] | 32.7 | 32.4 | 31.4 | 18.9 | 19.9 | 19.9 | 15.7 | 15.4 | 16.0 | 15.2 | 14.5 | 14.8 |
| C | PoseSelector [4] | 27.1 | 27.7 | 26.9 | 14.3 | 14.6 | 14.0 | 15.7 | 15.7 | 16.0 | 15.3 | 15.0 | 15.1 |
| | PCA-RI [3] | 31.5 | 30.4 | 31.1 | 19.6 | 19.5 | 19.7 | 16.8 | 16.7 | 16.8 | 14.6 | 14.8 | 14.8 |
| D | EOMP [5] | 34.2 | 34.3 | 34.0 | 28.5 | 28.3 | 28.0 | 25.2 | 25.0 | 24.8 | 16.5 | 16.5 | 16.7 |
| | DLAN [2] | 58.8 | 58.7 | 58.7 | 42.8 | 44.0 | 42.9 | 46.7 | 46.7 | 46.8 | 40.9 | 40.9 | 40.2 |
| | RIPT (ours) | **64.7** | **64.8** | **64.8** | **49.3** | **49.0** | **49.4** | **47.0** | **47.3** | **47.3** | **41.4** | **42.0** | **41.7** |

Table 3 shows retrieval accuracy of the features learned by different RI DNNs using (a restricted version of) SDMM. In the table, DNNs belonging to the same group employ the same approach to achieve rotation invariance. Specifically, the group A extracts inherently RI local feature as input to the DNN, the group B adopts rotation-equivariant layers, the group C normalizes rotation of global 3D shapes, and the group D normalizes rotation of local 3D shapes.

An important finding of this experiment is that *an RI DNN designed for supervised learning does not necessarily learn accurate RI features under an SSL scenario*. All the existing DNNs in Table 3 exhibit rotation invariance, but their retrieval accuracies are significantly lower than RIPT. As mentioned in Sections 1 and 2, discarding pose information of local regions hinders feature learning by several DNNs in the group A and D. The DNNs in the group B, which employ rotation equivariant layers, do not suffer from the pose information loss, but their feature accuracies are low due to the strong constraints (e.g., linearity) imposed on the feature transformation layers. The DNNs in the group C, which attempt to normalize the rotation of an entire 3D point set, face a difficulty in normalizing the rotation especially of realistic 3D point sets having background and noise. Among the competitor DNNs, DLAN performs the best. This is because DLAN randomizes the scale of regions sampled from an input 3D point set and describes these regions by using 3D grid-structured features. Therefore, DLAN can process, to some extent, global-scale 3D grid features as RIPT does.

Table 4 compares computation costs required in the experiment of Table 3. In Table 4, all the DNN architectures are trained by using the restricted version of SDMM which omits mixed views. During training, the minibatch size for all the DNNs is fixed at $E = 32$, meaning that each DNN processes a minibatch containing $32 \times 4 = 128$ 3D point sets including two global views and two local views per training sample. We ran the codes on two GPUs with a total of 48 GBytes VRAM. To measure GPU memory consumption during training, we use the *NVIDIA System Management Interface* implemented as the *nvidia-smi* command. To measure computation time per training epoch, we measure wall-clock time between the beginning of one epoch and the beginning of the next epoch.

Our SDMM is a memory-intensive SSL algorithm since it creates a minibatch containing a large number of samples and requires two DNNs, i.e., student and teacher, having an identical architecture. Nevertheless, Table 4 shows that our RIPT requires only 7.2 GBytes of VRAM and less than 100 seconds per epoch for the training on the MN40 dataset. The high efficiency of RIPT stems from its tokenization and feature refinement processes. RI-Tokenizer converts an input 3D point set into relatively small number (i.e., $T = 256$) of token features and TS-Transformer refines them by using only two self-attention blocks. Such a lightweight DNN architecture is potentially useful for SSL on large-scale datasets (e.g., [10]) composed of numerous 3D point sets.

Table 4. Comparison of training efficiency of RI DNN architectures using SDMM.

| Group | DNN architectures | Number of DNN parameters | GPU memory consumption during training [Bytes] | Computation time [s] per training epoch (MN40 dataset) |
|---|---|---|---|---|
| A | LGR-Net [24] | 5.5M | 30.4G | 232 |
| A | RIConv++ [9] | 0.6M | 15.6G | 303 |
| A | PaRI-Conv [8] | 1.1M | 39.8G | 470 |
| B | VN-PointNet [28] | **0.4M** | 23.2G | 170 |
| B | VN-DGCNN [28] | 0.7M | 40.8G | 339 |
| B | REQNN [27] | 1.1M | 38.0G | 233 |
| C | PoseSelector [4] | 2.8M | 41.3G | 252 |
| C | PCA-RI [3] | 1.2M | 46.0G | 421 |
| D | EOMP [5] | 1.8M | 40.5G | 235 |
| D | DLAN [2] | 31.7M | 11.4G | 626 |
|  | RIPT (ours) | 6.0M | **7.2G** | **96** |

**Evaluation on larger-scale datasets**: Table 5 compares the retrieval accuracy on the SN55 and SO-PB datasets. This experiment evaluates MacroMAP [%] only under the Nr/Rr setting since the rotation invariance of the algorithms in Table 5 is demonstrated in Table 2 and Table 3. Similar to the results in Tables 2 and 3, our proposed algorithm outperforms the existing ones even on the larger-scale datasets consisting of both synthetic and realistic 3D point sets.

Table 5. Retrieval accuracy (macroMAP [%]) of RI features learned on larger-scale datasets.

| Algorithms | SN55 dataset | SO-PB dataset |
|---|---|---|
| PPF-FoldNet [45] | 14.8 | 11.6 |
| RIConv++ [9] + SDMM | 17.9 | 21.2 |
| VN-PointNet [28] + SDMM | 17.2 | 11.5 |
| EOMP [5] + SDMM | 23.1 | 19.1 |
| DLAN [2] + SDMM | 28.7 | 30.0 |
| RIPT + SDMM (ours) | **34.1** | **34.6** |

**Evaluation under classification and clustering scenarios**: Table 6 compares the accuracies for classification and clustering with the Nr/Rr rotation setting. Table 6 exhibits a similar tendency to the abovementioned retrieval experiments. That is, the proposed RIPT + SDMM outperforms the other algorithms both in the classification and clustering tasks. This result suggests that the object-level 3D point set features learned by our algorithm have high generalizability and can be applied to diverse analysis tasks that involve object-level 3D point set features.

Table 6. Classification accuracy (macro-averaged accuracy [%]) and clustering accuracy (NMI) of RI features.

| Algorithms | Classification | | | | Clustering | | | |
|---|---|---|---|---|---|---|---|---|
|  | MN10 | MN40 | SO-OO | SO-OB | MN10 | MN40 | SO-OO | SO-OB |
| PPF-FoldNet [45] | 72.1 | 67.0 | 45.3 | 34.9 | 0.425 | 0.514 | 0.292 | 0.254 |
| RIConv++ [9] + SDMM | 78.0 | 65.2 | 60.0 | 57.4 | 0.436 | 0.573 | 0.387 | 0.391 |
| VN-PointNet [28] + SDMM | 77.2 | 70.0 | 45.3 | 36.9 | 0.362 | 0.513 | 0.283 | 0.245 |
| EOMP [5] + SDMM | 69.7 | 66.3 | 51.2 | 43.1 | 0.429 | 0.581 | 0.406 | 0.370 |
| DLAN [2] + SDMM | 89.2 | 78.9 | 73.9 | 68.8 | 0.682 | 0.719 | 0.586 | 0.547 |
| RIPT + SDMM (ours) | **90.8** | **83.0** | **77.0** | **73.7** | **0.791** | **0.757** | **0.599** | **0.579** |

### 4.2.2. In-depth evaluation of the proposed algorithm

This section provides a more comprehensive evaluation of our proposed algorithm. We evaluate each of the three components of our algorithm: RI-Tokenizer, TS-Transformer, and SDMM.

**RI-Tokenizer:** The tokenizer converts an input 3D point set into multiple global-scale tokens. To verify the efficacy of using global-scale tokens, we introduce a hyperparameter $s$ that controls the scale of token features. For each token point **c** chosen from $N$ 3D points per shape, we find $\lfloor sN \rfloor$ neighboring 3D points of **c** to construct a region for tokenization. Setting $s = 1$ constructs global-scale regions. Figure 4 plots feature accuracies against the scale parameter $s$. Evidently, using large-scale regions (e.g., $s > 0.6$) contributes high retrieval accuracy for all the datasets we have experimented with. Our 3D grid features computed at large-scale mitigate the pose information loss problem. In contrast, using small-scale regions (e.g., $s < 0.1$) suffers from the pose information loss since each token fails to describe the spatial layout of the parts composing a 3D point set.

Figure 5 shows the relationship between the number of tokens $T$ and retrieval accuracy. Setting $T$ at larger than 100 help RIPT learn reasonably accurate features. Reducing $T$ below 100 significantly decreases retrieval accuracy, especially in the SO-OO and SO-OB datasets having realistic 3D point sets. This is probably because FPS prioritizes choosing background and noise 3D points as token points, which hampers the subsequent rotation normalization and feature extraction processes.

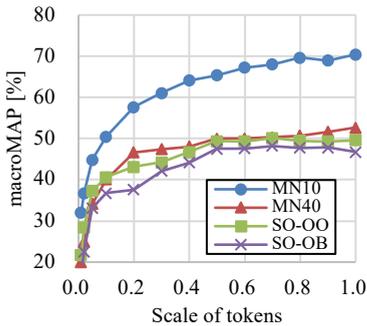
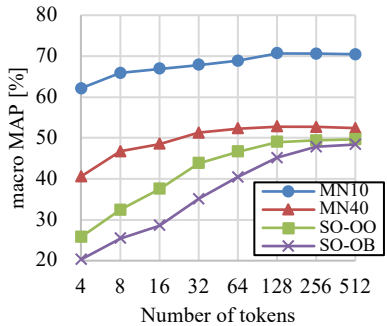
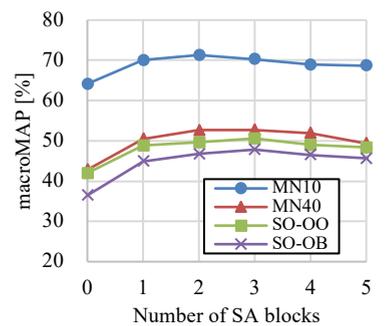

Figure 4. Retrieval accuracy plotted against the scale $s$ of tokens.

Figure 5. Retrieval accuracy plotted against the number $T$ of tokens.

Figure 6. Retrieval accuracy plotted against the number of self-attention (SA) blocks.

**TS-Transformer:** The transformer refines token sets by using the two SA blocks. Figure 6 plots retrieval accuracy against the number of SA blocks. We can observe that a peak of accuracy appears at two or three SA blocks for all the four datasets. Existing Transformer-based DNNs for 3D point sets require more blocks (e.g., four [21] and five [15]) to gradually refine local features into a global feature. Our TS-Transformer requires fewer layers for feature refinement since the input already represents global features. The use of small number of SA blocks contributes to the high efficiency shown in Table 4.

Table 7 demonstrates the effectiveness of the vector self-attention used in TS-Transformer. EdgeConv [20] is a type of graph convolution. Scalar self-attention [14] assigns a single weight to each token feature vector, while vector self-attention [15] assigns weights to every feature channel of a token feature vector. No attention means that the weight values are fixed at 1 regardless of input token features. Table 7 shows that the vector self-attention outperforms the other feature refinement layers. The improvement in accuracy of the vector self-attention over the scalar self-attention is particularly significant in the SO-OB dataset. We suspect that the vector self-attention is especially useful in suppressing feature channels that correspond to background and noise 3D points.

Table 7. Effectiveness of self-attention mechanism (macroMAP [%]).

| Feature refinement layers | MN10 dataset | MN40 dataset | SO-OO dataset | SO-OB dataset |
|---|---|---|---|---|
| EdgeConv [20] | 69.5 | 50.4 | 48.3 | 43.7 |
| No attention | 67.4 | 50.6 | 45.2 | 40.6 |
| Scalar self-attention [14] | 68.0 | 50.1 | 45.7 | 41.6 |
| Vector self-attention [15] | **70.5** | **52.0** | **49.8** | **47.3** |

In addition, we evaluate the feature refinement capability of the proposed TS-Transformer by comparing it against various feature refinement DNNs designed for 3D point sets. Specifically, we utilize PointNet [19], PointNet++ [70], DGCNN [20], PointMLP [71], and Transformer used in [47] and [48] as feature refinement DNNs. In all cases, the proposed RI-Tokenizer is employed to convert an input 3D point set into a token set $S$ defined in Section 3.2.1. The token set is then fed into one of the feature refinement DNNs. To eliminate the influence of self-distillation introduced by SDMM, this experiment employs the SimCLR algorithm [40] for self-supervised learning. We use the Nr/Rr setting.

Table 8 demonstrates that the proposed TS-Transformer significantly contributes to the acquisition of accurate RI point set features. Among the competitors, relatively shallow DNNs (PointNet, PointNet++, and DGCNN) perform better than deeper DNNs (PointMLP and Transformer). Too many layers/blocks are not suitable for refining the global-scale token features produced by RI-Tokenizer, as also shown in Figure 6. The results in Table 8 validate the design of our TS-Transformer which incorporates only a few self-attention blocks for feature refinement.

Table 8. Effectiveness of the proposed TS-Transformer (macroMAP [%]).

| Feature refinement DNN | MN10 dataset | MN40 dataset | SO-OO dataset | SO-OB dataset |
|---|---|---|---|---|
| PointNet [19] | 52.1 | 37.9 | 33.2 | 25.8 |
| PointNet++ [70] | 48.5 | 35.8 | 32.1 | 25.3 |
| DGCNN [20] | 51.8 | 38.9 | 35.2 | 25.0 |
| PointMLP [71] | 43.2 | 33.1 | 22.5 | 18.0 |
| Transformer used in [47] and [48] | 43.5 | 34.9 | 27.5 | 22.1 |
| TS-Transformer (proposed) | **59.0** | **40.5** | **36.4** | **31.5** |

**SDMM:** The novelty of SDMM lies in the simultaneous use of multi-crop and cut-mix data augmentation to create minibatches including highly diverse 3D point sets. Table 9 demonstrates the importance of combining the three types of views created by our data augmentation. Omitting local views and/or mixed views from SDMM significantly degrades feature accuracy. We suspect that both local and mixed views regularize the training and facilitate convergence to a better latent feature space. The local view would be particularly useful to help RIPT learn local-to-global correspondences of 3D shapes. The mixed view seems beneficial also for learning a smooth latent feature space where 3D shape features belonging to different semantic categories are continuously distributed based on their shape similarity.

To verify the effectiveness of SDMM, we compare accuracies of features extracted by RIPT when it is trained by using different SSL algorithms. Table 10 shows that the two SSL algorithms which employ self-distillation, i.e., BYOL and our SDMM, outperform the other SSL algorithms. One possible explanation for the success of self-distillation is that it performs as a form of curriculum learning [68]. That is, the teacher DNN can generate pseudo-labels that are adapted to the training progress of the student DNN. We speculate that self-distillation facilitates convergence to a better solution, which is one of the benefits of curriculum learning [68], resulting in accurate 3D point set features. Among the two self-distillation based approaches, SDMM outperforms BYOL since SDMM employs multi-crop and cut-mix data augmentation for diverse training samples while BYOL does not.

Table 9. Effectiveness of multi-crop and cut-mix data augmentation in SDMM (macroMAP [%]).

| Global view | Local view | Mixed view | MN10 dataset | MN40 dataset | SO-OO dataset | SO-OB dataset |
|---|---|---|---|---|---|---|
| Yes | No | No | 59.9 | 30.3 | 35.7 | 27.2 |
| Yes | Yes | No | 64.8 | 49.0 | 47.3 | 42.0 |
| Yes | No | Yes | 68.7 | 48.1 | 43.0 | 41.1 |
| Yes | Yes | Yes | **70.5** | **52.0** | **49.8** | **47.3** |

Table 10. Impact of SSL algorithms on 3D point set feature learning by RIPT (macroMAP [%]).

| SSL algorithms | MN10 dataset | MN40 dataset | SO-OO dataset | SO-OB dataset |
|---|---|---|---|---|
| Instance Discrimination [32] | 56.4 | 36.0 | 34.1 | 28.2 |
| DeepCluster [38] | 53.2 | 35.5 | 32.6 | 27.7 |
| SimCLR [40] | 59.0 | 40.5 | 36.4 | 31.5 |
| MoCo [41] | 55.9 | 39.8 | 37.3 | 31.0 |
| SimSiam [42] | 56.6 | 34.1 | 36.0 | 27.7 |
| BYOL [34] | 60.1 | 42.9 | 44.0 | 34.4 |
| SDMM (ours) | **70.5** | **52.0** | **49.8** | **47.3** |

## 5. Conclusion and future work

Despite the importance of Self-Supervised Learning (SSL) of Rotation-Invariant (RI) 3D point set features, it has not been sufficiently studied. To obtain RI and accurate object-level 3D point set features, we proposed a novel DNN architecture called *Rotation-Invariant Point set token Transformer* (*RIPT*), and its SSL algorithm called *Self-Distillation with Multi-crop and cut-Mix point set augmentation* (*SDMM*). Conventional DNNs having rotation invariance face the pose loss problem [8]. RIPT addresses this by decomposing an input 3D point set into highly overlapping multiple *global* scale regions, or tokens. A set of global tokens is then transformed into an expressive RI latent 3D shape feature via feature refinement layers using a self-attention mechanism. SDMM effectively trains the student RIPT by leveraging pseudo-labels generated by the teacher RIPT. The 3D point sets for training are diversified by using the multi-crop and cut-mix data augmentation techniques. Through the comprehensive experimental evaluation, we found that:

- Existing RI DNN architectures designed for supervised learning do not necessarily work well under the SSL scenario. To the best of our knowledge, this paper is the first to point out this issue.

- RIPT trained by SDMM can extract RI 3D point set features more accurate than the existing algorithms we have experimented. In addition, RIPT exhibits higher computational efficiency compared to the existing RI DNNs.

- The key components of our proposed algorithm, i.e., global-scale tokenization, feature refinement using self-attention, and SSL combining self-distillation and multi-crop/cut-mix data augmentation, all contribute to the high accuracy of the learned 3D point set features.

As future work, there are two possible directions, that are, improving the feature accuracy and increasing the versatility of our algorithm. Although our algorithm achieves higher feature accuracy than existing algorithms, it is still not sufficient for practical use. Training on larger 3D shape datasets, such as Objaverse constructed by Deitke et al. [10], may lead to higher feature accuracy. In addition, it is worth exploring the use of new SSL frameworks developed primarily in the fields of 2D image analysis and natural language processing. In terms of versatility, the applications of our current algorithm are limited to, for example, retrieval, clustering, and classification of 3D point sets. We will consider modifying the architecture of RIPT to make it applicable to segmentation and registration of 3D point sets having inconsistent orientations.

## Acknowledgements

Takahiko Furuya and Zhoujie Chen contributed equally. This work was supported by the Japan Society for the Promotion of Science (JSPS) KAKENHI (Grant No. 21K17763) and was supported by Zhejiang Provincial Natural Science Foundation of China (Grant No. LY22F020028).

# References

[1] Y. Guo, H. Wang, Q. Hu, H. Liu, L. Liu, and M. Bennamoun, Deep Learning for 3D Point Clouds: A Survey, in TPAMI, Vol. 43, No. 12, pp. 4338–4364, 2021.

[2] T. Furuya and R. Ohbuchi, Deep Aggregation of Local 3D Geometric Features for 3D Model Retrieval, in BMVC 2016, Vol. 7, p. 8, 2016.

[3] Z. Xiao, H. Lin, R. Li, L. Geng, H. Chao, and S. Ding, Endowing Deep 3d Models With Rotation Invariance Based On Principal Component Analysis, in ICME 2020, pp. 1-6, 2020.

[4] F. Li, K. Fujiwara, F. Okura, and Y. Matsushita, A Closer Look at Rotation-Invariant Deep Point Cloud Analysis, in ICCV 2021, pp. 16218-16227, 2021.

[5] S. Luo, J. Li, J. Guan, Y. Su, C. Cheng, J. Peng, and J. Ma, Equivariant Point Cloud Analysis via Learning Orientations for Message Passing, in CVPR 2022, pp. 18910-18919, 2022.

[6] C. Chen, G. Li, R. Xu, T. Chen, M. Wang and L. Lin, ClusterNet: Deep Hierarchical Cluster Network With Rigorously Rotation-Invariant Representation for Point Cloud Analysis, in CVPR 2019, pp. 4989-4997, 2019.

[7] Z. Zhang, B.-S. Hua, D. W. Rosen, and S.-K. Yeung, Rotation Invariant Convolutions for 3D Point Clouds Deep Learning, in 3DV 2019, pp. 204-213, 2019.

[8] R. Chen and Y. Cong, The Devil is in the Pose: Ambiguity-free 3D Rotation-invariant Learning via Pose-aware Convolution, in CVPR 2022, pp. 7462-7471, 2022.

[9] Z. Zhang, B.-S. Hua, and S.-K. Yeung, RIConv++: Effective Rotation Invariant Convolutions for 3D Point Clouds Deep Learning, in IJCV, vol. 130, pp. 1228–1243, 2022.

[10] M. Deitke, D. Schwenk, J. Salvador, L. Weihs, O. Michel, E. VanderBilt, L. Schmidt, K. Ehsani, A. Kembhavi, A. Farhadi, Objaverse: A Universe of Annotated 3D Objects, arXiv preprint, arXiv:2212.08051, 2022.

[11] A. Xiao, J. Huang, D. Guan, X. Zhang, S. Lu, and L. Shao, Unsupervised Point Cloud Representation Learning with Deep Neural Networks: A Survey, in TPAMI, 2023.

[12] H. Deng, T. Birdal, and S. Ilic, PPF-FoldNet: Unsupervised Learning of Rotation Invariant 3D Local Descriptors, in ECCV 2018, pp. 620–638, 2018.

[13] M. Marcon, R. Spezialetti, S. Salti, L. Silva, and L. D. Stefano, Unsupervised Learning of Local Equivariant Descriptors for Point Clouds, in TPAMI, Vol. 44, No. 12, pp. 9687-9702, 2022.

[14] A. Vaswani, N. Shazeer, N. Parmar, J. Uszkoreit, L. Jones, A. N. Gomez, Ł. Kaiser, and I. Polosukhin, Attention is All you Need, in NeurIPS 30, 2017.

[15] H. Zhao, L. Jiang, J. Jia, P. H.S. Torr, and V. Koltun, Point Transformer, in ICCV 2021, pp. 16259-16268, 2021.

[16] M. Caron, H. Touvron, I. Misra, H. Jégou, J. Mairal, P. Bojanowski, and A. Joulin, Emerging Properties in Self-Supervised Vision Transformers, in ICCV 2021, pp. 9650-9660, 2021.

[17] S. Yun, D. Han, S. Chun, S. Joon Oh, Y. Yoo, and J. Choe, CutMix: Regularization Strategy to Train Strong Classifiers with Localizable Features, in ICCV 2019, pp. 6022-6031, 2019.

[18] M. Caron, I. Misra, J. Mairal, and P. Goyal, P. Bojanowski, and A. Joulin, Unsupervised Learning of Visual Features by Contrasting Cluster Assignments, in NeurIPS 33, 2020.

[19] C. R. Qi, H. Su, M. Kaichun, and L. J. Guibas, PointNet: Deep Learning on Point Sets for 3D Classification and Segmentation, in CVPR 2017, pp. 77-85, 2017.

[20] Y. Wang, Y. Sun, Z. Liu, S. E. Sarma, M. M. Bronstein, and J. M. Solomon, Dynamic Graph CNN for Learning on Point Clouds, in ACM ToG, Vol. 38, No. 5, Article No. 146, 2019.

[21] M.-H. Guo, J.-X. Cai, Z.-N. Liu, T.-J. Mu, R. R. Martin, and S.-M. Hu, PCT: Point cloud transformer, in Computational Visual Media, Vol. 7, No. 2, pp. 187-199, 2021.

[22] S. Kim, J. Park, and B. Han, Rotation-Invariant Local-to-Global Representation Learning for 3D Point Cloud, in NeurIPS 33, 2020

[23] J. Zhang, M.-Y. Yu, R. Vasudevan, and M. Johnson-Roberson, Learning Rotation-Invariant Representations of Point Clouds Using Aligned Edge Convolutional Neural Networks, in 3DV, pp. 200-209, 2020.

[24] C. Zhao, J. Yang, X. Xiong, A. Zhu, Z. Cao, and X. Li, Rotation Invariant Point Cloud Analysis: Where Local Geometry Meets Global Topology, in Pattern Recognition, vol. 127, 2022.

[25] X. Li, R. Li, G. Chen, C. Fu, D. Cohen-Or, and P. Heng, A Rotation-Invariant Framework for Deep Point Cloud Analysis, in TVCG, Vol. 28, No. 12, pp. 4503-4514, 2022.

[26] X. Sun, Z. Lian, and J. Xiao, SRINet: Learning Strictly Rotation-Invariant Representations for Point Cloud Classification and Segmentation, in MM 2019, pp. 980–988, 2019.

[27] W. Shen, B. Zhang, S. Huang, Z. Wei, and Q. Zhang, 3D-Rotation-Equivariant Quaternion Neural Networks, in ECCV 2020, pp. 531–547, 2020.

[28] C. Deng, O. Litany, Y. Duan, A. Poulenard, A. Tagliasacchi, and L. Guibas, Vector Neurons: A General Framework for SO(3)-Equivariant Networks, in ICCV 2021, pp. 12180-12189, 2021.

[29] S. Assaad, C. Downey, R. Al-Rfou, N. Nayakanti, and B. Sapp, VN-Transformer: Rotation-Equivariant Attention for Vector Neurons, in TMLR, 2023.

[30] L. Ericsson, H. Gouk, C. C. Loy, and T. M. Hospedales, Self-Supervised Representation Learning: Introduction, advances, and challenges, in IEEE Signal Processing Magazine, Vol. 39, No. 3, pp. 42-62, 2022.

[31] G. E. Hinton and R. R. Salakhutdinov, Reducing the Dimensionality of Data with Neural Networks, in Science, Vol. 313, Issue 5786, pp. 504–507, 2006.

[32] A. Dosovitskiy, J. T. Springenberg, M. Riedmiller, and T. Brox, Discriminative Unsupervised Feature Learning with Convolutional Neural Networks, *Proc. NIPS 2014*, pp. 766–774, 2014.

[33] M. Ye, X. Zhang, P. C. Yuen, and S.-F. Chang, Unsupervised Embedding Learning via Invariant and Spreading Instance Feature, Proc. CVPR 2019, pp. 6210–6219, 2019.

[34] J.-B. Grill, F. Strub, F. Altché, C. Tallec, P. Richemond, E. Buchatskaya, C. Doersch, B. A. Pires, Z. Guo, M. G. Azar, B. Piot, K. Kavukcuoglu, R. Munos, and M. Valko, Bootstrap Your Own Latent: A New Approach to Self-Supervised Learning, in NeurIPS 33, pp. 21271-21284, 2020.

[35] K. He, X. Chen, S. Xie, Y. Li, P. Dollár, and R. Girshick, Masked Autoencoders Are Scalable Vision Learners, in CVPR 2022, pp. 16000-16009, 2022.

[36] A. Dosovitskiy, L. Beyer, A. Kolesnikov, D. Weissenborn, X. Zhai, T. Unterthiner, M. Dehghani, M. Minderer, G. Heigold, S. Gelly, J. Uszkoreit, N. Houlsby, An Image is Worth 16x16 Words: Transformers for Image Recognition at Scale, in ICLR 2021, 2021.

[37] Z. Wu, Y. Xiong, S. X. Yu, and D. Lin, Unsupervised Feature Learning via Non-parametric Instance Discrimination, in CVPR 2018, pp. 733–3742, 2018.

[38] M. Caron, P. Bojanowski, A. Joulin, and M. Douze, Deep Clustering for Unsupervised Learning of Visual Features, in ECCV 2018, pp. 132–149, 2018.

[39] C. Zhuang, A. L. Zhai, and D. Yamins, Local Aggregation for Unsupervised Learning of Visual Embeddings, in ICCV 2019, pp. 6002–6012, 2019.

[40] T. Chen, S. Kornblith, M. Norouzi, and G. Hinton, A Simple Framework for Contrastive Learning of Visual Representations, in ICMR 2020, pp. 1597–1607, 2020.

[41] K. He, H. Fan, Y. Wu, S. Xie, and R. Girshick, Momentum Contrast for Unsupervised Visual Representation Learning, in CVPR 2020, pp. 9729–9738, 2020.

[42] X. Chen and K. He, Exploring Simple Siamese Representation Learning, in CVPR 2021, pp. 15750-15758, 2021.

[43] G. Hinton, O. Vinyals, and J. Dean, Distilling the Knowledge in a Neural Network, in NIPS Deep Learning and Representation Learning Workshop, 2015.

[44] S. Ren, H. Wang, Z. Gao, S. He, A. Yuille, Y. Zhou, and C. Xie, A Simple Data Mixing Prior for Improving Self-Supervised Learning, in CVPR 2022, pp. 14575-14584, 2022.

[45] Y. Yang, C. Feng, Y. Shen, and D. Tian, FoldingNet: Point Cloud Auto-Encoder via Deep Grid Deformation, in CVPR 2018, pp. 206-215, 2018.

[46] W. Sun, A. Tagliasacchi, B. Deng, S. Sabour, S. Yazdani, G. E. Hinton, and K. M. Yi, Canonical Capsules: Self-supervised Capsules in Canonical Pose, in NeurIPS 34, pp. 24993-25005, 2021.

[47] X. Yu, L. Tang, Y. Rao, T. Huang, J. Zhou, and J. Lu, Point-BERT: Pre-Training 3D Point Cloud Transformers with Masked Point Modeling, in CVPR 2022, pp. 19313-19322, 2022.

[48] H. Liu, M. Cai, and Y. J. Lee, Masked Discrimination for Self-Supervised Learning on Point Clouds, in ECCV 2022, pp. 657-675, 2022.

[49] P.-S. Wang, Y.-Q. Yang, Q.-F. Zou, Z. Wu, Y. Liu, and X. Tong, Unsupervised 3D Learning for Shape Analysis via Multiresolution Instance Discrimination, in AAAI 2021, Vol. 35, No. 4, pp. 2773-2781, 2021.

[50] K. Fu, P. Gao, R. Zhang, H. Li, Y. Qiao, and M. Wang, Distillation with Contrast is All You Need for Self-Supervised Point Cloud Representation Learning, arXiv preprint, arXiv:2202.04241, 2022.

[51] A. Sanghi, Info3D: Representation Learning on 3D Objects using Mutual Information Maximization and Contrastive Learning, in ECCV 2020, pp. 626-642, 2020.

[52] B. Du, X. Gao, W. Hu, and X. Li, Self-Contrastive Learning with Hard Negative Sampling for Self-supervised Point Cloud Learning, in ACM Multimedia 2021, pp. 3133-3142, 2021.

[53] T. Furuya and R. Ohbuchi, DeepDiffusion: Unsupervised Learning of Retrieval-Adapted Representations via Diffusion-Based Ranking on Latent Feature Manifold, in IEEE Access, Vol. 10, pp. 116287-116301, 2022.

[54] T. Furuya and R. Ohbuchi, Diffusion-on-Manifold Aggregation of Local Features for Shape-based 3D Model Retrieval, in ICMR 2015, pp. 171–178, 2015.

[55] J. Zhang, L. Chen, B. Ouyang, B. Liu, J. Zhu, Y. Chen, Y. Meng, and D. Wu, PointCutMix: Regularization Strategy for Point Cloud Classification, in Neurocomputing, Vol. 505, pp. 58-67, 2022.
[56] Y. Eldar, M. Lindenbaum, M. Porat, and Y. Y. Zeevi, The Farthest Point Strategy for Progressive Image Sampling, in TIP, Vol. 6, No. 9, pp. 1305-1315, 1997.
[57] S. Salti, F. Tombari, and L. D. Stefano, SHOT: Unique Signatures of Histograms for Surface and Texture Description, in CVIU, Vol. 125, pp. 251-264, 2014.
[58] K. He, X. Zhang, S. Ren, and J. Sun, Deep Residual Learning for Image Recognition, in CVPR 2016, pp. 770-778, 2016.
[59] S. Ioffe and C. Szegedy, Batch Normalization: Accelerating Deep Network Training by Reducing Internal Covariate Shift, in ICML 2015, Vol. 37, pp. 448-456, 2015.
[60] D. Hendrycks and K. Gimpel, Gaussian Error Linear Units (GELUs), in arXiv preprint, arXiv:1606.08415, 2016.
[61] D. P. Kingma and J. Ba, Adam: A Method for Stochastic Optimization, in ICLR 2015, 2015.
[62] P. Goyal, P. Dollár, R. Girshick, P. Noordhuis, L. Wesolowski, A. Kyrola, A. Tulloch, Y. Jia, K. He, Accurate, Large Minibatch SGD: Training ImageNet in 1 Hour, in arXiv preprint, arXiv:1706.02677, 2017.
[63] Z. Wu, S. Song, A. Khosla, F. Yu, L. Zhang, X. Tang, and J. Xiao, 3D ShapeNets: A Deep Representation for Volumetric Shapes, in CVPR 2015, pp. 1912-1920, 2015.
[64] M. A. Uy, Q.-H. Pham, B.-S. Hua, T. Nguyen, and S.-K. Yeung, Revisiting Point Cloud Classification: A New Benchmark Dataset and Classification Model on Real-World Data, in ICCV 2019, pp. 1588-1597, 2019.
[65] A. X. Chang, T. Funkhouser, L. Guibas, P. Hanrahan, Q. Huang, Z. Li, S. Savarese, M. Savva, S. Song, H. Su, J. Xiao, L. Yi, and F. Yu, ShapeNet: An Information-Rich 3D Model Repository, in arXiv preprint, arXiv:1512.03012, 2015.
[66] R. Ohbuchi, T. Minamitani, and T. Takei, Shape-Similarity Search of 3D Models by using Enhanced Shape Functions, in IJCAT, Vol. 23, No. 2/3/4, pp. 70–85, 2005.
[67] A. Paszke, S. Gross, F. Massa, A. Lerer, J. Bradbury, G. Chanan, T. Killeen, Z. Lin, N. Gimelshein, L. Antiga, A. Desmaison, A. Köpf, E. Yang, Z. DeVito, M. Raison, A. Tejani, S. Chilamkurthy, B. Steiner, L. Fang, J. Bai, and S. Chintala, PyTorch: an imperative style, high-performance deep learning library, in NeurIPS 2019, Article No.: 721, pp. 8026–8037, 2019.
[68] Y. Bengio, J. Louradour, R. Collobert, and J. Weston, Curriculum Learning, in ICML 2009, pp. 41–48, 2009.
[69] B. E. Boser, I. M. Guyon, and V. N. Vapnik, A Training Algorithm for Optimal Margin Classifiers, in COLT '92, pp. 144–152, 1992.
[70] C. R. Qi, L. Yi, H. Su, and L. J. Guibas, PointNet++: Deep Hierarchical Feature Learning on Point Sets in a Metric Space, in NeurIPS 30, pp. 5099–5108, 2017.
[71] X. Ma, C. Qin, H. You, H. Ran, Y. Fu, Rethinking Network Design and Local Geometry in Point Cloud: A Simple Residual MLP Framework, in ICLR 2022, 2022.
[72] Y. Chen, V. T. Hu, E. Gavves, T. Mensink, P. Mettes, P. Yang, and C. G. M. Snoek, PointMixup: Augmentation for Point Clouds, in ECCV 2020 pp. 330–345, 2020.